\documentclass[times,twocolumn,final]{elsarticle}
\usepackage{natbib}
\usepackage{url}
\usepackage{amsmath,amssymb,amsfonts}
\usepackage{multicol}
\usepackage{graphics}
\usepackage[table]{xcolor}
\usepackage{colortbl}
\usepackage{multirow}
\usepackage{multicol}
\usepackage{booktabs}
\journal{Neurocomputing}

\biboptions{authoryear}

\begin{document}

\begin{frontmatter}

\title{SRPL-SFDA: SAM-Guided Reliable Pseudo-Labels  for Source-Free Domain Adaptation in Medical Image Segmentation}
\author[1]{Xinya Liu}
\author[1,2]{Jianghao Wu}
\author[3]{Tao Lu\corref{mycorrespondingauthor}}
\author[1,2]{Shaoting Zhang}
\author[1,2]{Guotai~Wang\corref{mycorrespondingauthor}}
\address[1]{School of Mechanical and Electrical Engineering, University of Electronic Science and Technology of China, Chengdu, China}
\address[2]{Shanghai Artificial Intelligence Laboratory, Shanghai, China}
\address[3]{The Department of Radiology, Sichuan Provincial People’s Hospital, University of Electronic Science and Technology of China, Chengdu, China}

\cortext[mycorrespondingauthor]{Corresponding authors}
\ead{guotai.wang@uestc.edu.cn}

\begin{abstract}
Domain Adaptation (DA) is crucial for robust deployment of medical image segmentation models when applied to new clinical centers with significant domain shifts. Source-Free Domain Adaptation (SFDA) is appealing as it can deal with privacy concerns and access constraints on source-domain data during adaptation to target-domain data. However, SFDA faces challenges such as insufficient supervision in the target domain with unlabeled images. In this work, we propose a Segment Anything Model (SAM)-guided Reliable Pseudo-Labels method for SFDA (SRPL-SFDA) with three key components: 1) Test-Time Tri-branch Intensity Enhancement (T3IE) that not only improves quality of raw pseudo-labels in the target domain, but also leads to SAM-compatible inputs with three channels to better leverage SAM's zero-shot inference ability for refining the pseudo-labels; 2) A reliable pseudo-label selection module that rejects low-quality pseudo-labels based on Consistency of Multiple SAM Outputs (CMSO) under input perturbations with T3IE; and 3) A reliability-aware training procedure in the unlabeled target domain where reliable pseudo-labels are used for supervision and unreliable parts are regularized by entropy minimization. Experiments conducted on two multi-domain medical image segmentation datasets for fetal brain and the prostate respectively demonstrate that: 1) SRPL-SFDA effectively enhances pseudo-label quality in the unlabeled target domain, and improves SFDA performance by leveraging the reliability-aware training; 2) SRPL-SFDA outperformed state-of-the-art SFDA methods, and its performance is close to that of supervised training in the target domain. The code of this work is available online: \url{https://github.com/HiLab-git/SRPL-SFDA}.
\end{abstract}


\end{frontmatter}


\section{Introduction}
Medical image segmentation is important for disease diagnosis, treatment planning and guidance~\citep{wang2018deepigeos}. In recent years, deep learning has significantly advanced this field, enabling precise segmentation of various anatomical structures and lesions~\citep{ma2024segment}. This success primarily relies on the assumption that training and testing images come from the same distribution~\citep{zhang2024cluster,xia2024comprehensive}. However, in practical clinical applications, there are significant distribution differences (a.k.a., domain shifts) between training (source domain) and testing (target domain) datasets due to variations in scanners, imaging protocols, and image quality~\citep{wang2022towards}. Discrepancies between the source and target domains often lead to substantial performance degradation on unseen target domain datasets. Given the difficulty of covering all potential target domain images during the training phase~\citep{wagner2023post}, it is crucial to develop methods that can quickly adapt to new target domains post-deployment. 

Unsupervised Domain Adaptation (UDA) is an attractive method for addressing this challenge~\citep{chen2020unsupervised,han2021deep}, 
as it does not require annotations in the target domain for model adaptation, considering the time-consuming and laborious process of collecting labels. Existing UDA methods typically tackle this issue by aligning the appearance~\citep{zhu2017unpaired,park2020contrastive}, feature distribution~\citep{ge2023unsupervised,liu2022cada}, or output structure between the source and target domains~\citep{vu2019advent}. However, they often require simultaneous  access to source and target domain images for optimizing the alignment. Due to privacy and bandwidth concerns, this is not always feasible in clinical practice, making the UDA methods not applicable when source data are not accessible after model deployment~\citep{guan2021domain}.

To address this issue, Source-Free Domain Adaptation (SFDA) methods have shown great potential due to their ability to adapt a pre-trained model to a target domain dataset without accessing the source-domain data~\citep{liu2023memory,liang2021source,raychaudhuri2023prior,wen2023source}. 
Existing SFDA approaches have employed auxiliary tasks such as rotation prediction~\citep{sun2020test} and autoencoder-based image reconstruction~\citep{he2021autoencoder} to facilitate target domain adaptation. However, these methods are often constrained by the requirement of an extra branch for the auxiliary task, which limits their applicability to general segmentation models. To overcome these limitations, more flexible and general solutions have been proposed, making SFDA  independent of the pre-training process. For example, 
Test entropy minimization (TENT)~\citep{wang2020tent} and Adaptive Mutual Information (AdaMI)~\citep{bateson2022source} adapt the models by directly minimizing prediction entropy in the target domain. However, the supervisory power of entropy minimization is often inadequate, leading to a lot of over-confident but erroneous predictions in the target domain. Additionally, methods such as Uncertainty-aware Pseudo-Labeling (UPL-SFDA)~\citep{wu2023upl}, Denoised Pseudo-Labeling (DPL)~\citep{chen2021source}, and Category-Level Regularization (CLR)~\citep{xu2022denoising} focus on developing denoising strategies to generate reliable pseudo-labels for target domain data. Nonetheless, due to the large inter-domain distribution shift, the generated pseudo-labels are often quite unreliable~\citep{litrico2023guiding}, making it necessary to explore more effective methods to obtain high-quality pseudo-labels and learn from noise pseudo-labels. 

In this work, we propose a general and efficient method for generating and mining high-quality pseudo-labels for SFDA in medical image segmentation tasks. We introduce a reliable pseudo-labels refinement technique guided by the Segment Anything Model (SAM)~\citep{kirillov2023segment}. As SAM is effective for image segmentation with high generalizability based on prompt-based inference, it is natural to leverage SAM to obtain pseudo-labels for target domain images in SFDA by using the model's outputs as prompts. However, directly applying SAM in this scenario has several challenges that may limit the performance of SFDA. First, the source model usually has a poor performance on target domain images, which leads to inaccurate prompts for SAM to obtain good results. Second, as SAM is trained on natural images, the gap from natural images to medical images limits SAM's performance on the target domain. Thirdly, due to SAM's imperfect outputs on the target domain, directly using them as pseudo-labels to supervise the target domain model will lead to suboptimal results, therefore it is essential to identify reliable ones in SAM's outputs. We introduce a  Test-Time Tri-branch
Intensity Enhancement (T3IE) method for SAM's inference that helps to deal with these issues simultaneously, where the input image is augmented by histogram equalization, domain-level and image-level adaptive contrast adjustments. The ensemble of the  T3IE outputs not only leads to improved predictions of the source model for better prompts of SAM, but also obtains a three-channel input by concatenation that is more compatible for SAM, leading to better pseudo-labels. In addition, by sending each of the  T3IE outputs  to SAM, we obtain multiple pseudo-labels under input perturbation, where a consistency-based method can be applied to identify reliable pseudo-labels, which helps to better supervise the model for adaptation. 

The main contributions of this paper include: 
\begin{itemize}
    \item[1)] A novel SFDA method based on SAM-guided Reliable Pseudo-Labels (SRPL-SFDA) is proposed for medical image segmentation, which leverages SAM to refine pseudo-labels by using the raw predictions as prompts.
    \item[2)] To  reduce the gap between medical images and SAM's training set, we introduce a Test-Time Tri-branch Intensity Enhancement (T3IE)-based method that leads to better box prompts and more compatible inputs of SAM, which effectively improves pseudo-label quality. 
    \item[3)] A reliable pseudo-label mining method based on Consistency of Multiple SAM Outputs (CMSO) under input perturbations with T3IE is proposed, which effectively identifies high-quality pseudo-labels, and is combined with a Reliability-aware Pseudo-label Supervision and Regularization  (RPSR) loss for training the segmentation model in the target domain.  
\end{itemize}
This paper is a substantial extension of our previous work presented at ISBI-2024~\citep{liu2024rpl}, where a random intensity augmentation method is proposed to improve the quality of pseudo-labels that are combined with uncertainty estimation for noise-robust  training in the target domain. In this work, we further introduce SAM to refine the pseudo-labels, and replace the previous random intensity augmentation by SAM-compatible T3IE. Our method was evaluated on two multi-center datasets, including an MRI prostate segmentation dataset~\citep{liu2020ms} and an fetal brain segmentation dataset, demonstrating significant superiority over state-of-the-art SFDA methods.

\section{Related Works}
\subsection{Unsupervised Domain Adaptation}
UDA aims to transfer knowledge from a labeled source domain to an unlabeled target domain. Existing UDA methods primarily focus on three strategies: Firstly, input alignment methods adjust the raw data of the source and target domains to make them have similar appearance and intensity statistics~\citep{zhu2017unpaired,wolterink2017deep}. Secondly, feature alignment methods utilize specifically designed neural networks to learn domain-invariant features, minimizing the feature distribution distance between the source and target domains~\citep{hu2023dual,yang2023dc}. 
Thirdly, output alignment methods adapt the model in the target domain by using pseudo-labels or discriminators for supervising the target-domain model~\citep{xing2019adversarial,xing2020bidirectional}. Current UDA methods typically require full access to both source and target data. However, due to privacy protection requirements, the source-domain data is usually unavailable when the model is deployed to a new domain (e.g., a new medical center). In recent years, black-box domain adaptation has emerged as an important variant of UDA that can address privacy concerns by relying solely on the predictions of source models instead of the source data itself. Several works have explored black-box adaptation frameworks, such as \cite{liu2022unsupervised} used knowledge distillation with exponential mixup decay to adapt a black-box model to target domains. \cite{liuXiao2022unsupervised} integrated entropy minimization to improve target confidence. \cite{liang2022dine} proposed a DIstill and fine-tuNE (DINE) framework from single and multiple black-box predictors. Meanwhile, to more effectively address this issue, SFDA has emerged as an important solution that avoids potential violations of data privacy policies, and has garnered increasing attention in recent years~\citep{bateson2020source,liu2021source}.

\subsection{Source-Free Domain Adaptation}
SFDA adapts a pre-trained model to a target domain without access to source-domain images and annotations in the target domain. Current works~\citep{li2024comprehensive} primarily fall into three categories: data-driven, prediction regularization and pseudo-label-driven approaches. Data-driven methods concentrate on recalculating the statistics of target-domain images or using self-supervised learning tasks for adaptation. For instance, \cite{sun2020test} introduce an auxiliary branch for rotation angle prediction, where the model is optimized for the auxiliary prediction task in the target domain. However, the features for the auxiliary task may not match those for the segmentation task, which limits the performance for segmentation. \cite{nado2020evaluating} recalculate batch normalization statistics based on target-domain images, but the method is not applicable to models with other normalization techniques such as instance normalization, and cannot deal with more complex domain shifts. \cite{karani2021test} propose test-time adaptable networks that adapt an image normalization sub-network guided by implicit priors, improving robustness to scanner variations but relying on an independently trained denoising module. \cite{yang2022source}, on the other hand, leverages Fourier style mining to generate source-like images for domain alignment, achieving effective feature adaptation but requiring a two-stage process that increases computational complexity. Prediction regularization methods focus on adjusting parameters of certain layers based on unsupervised regularization losses. \cite{wang2020tent} update batch normalization parameters by minimizing prediction entropy in the target domain. \cite{bateson2022source} maximizes mutual information between target images and their label predictions, in addition to entropy minimization. However, these methods often under-perform due to insufficient constraints for learning. 

Pseudo-label-driven methods, on the other hand, generate and refine pseudo-labels in the target domain to guide the adaptation process, such as UPL-SFDA~\citep{wu2023upl}, DPL~\citep{chen2021source} and CLR~\citep{xu2022denoising}. UPL-SFDA leverages the growth of decoders in the target domain to obtain pseudo-labels with uncertainty estimation~\citep{wu2023upl}, yet its performance may be limited by the initial quality of pseudo-labels. DPL utilizes denoised pseudo-labeling to guide the adaptation process~\citep{chen2021source}, but the output of the denoising module may still contain mis-segmentation. CLR integrates uncertainty-rectified denoising with adaptive thresholding to refine pseudo-labels~\citep{xu2022denoising}, while it requires significant computational resources and meticulous hyper-parameter tuning. In addition, Confidence-Constrained Mean Teacher (CCMT)~\citep{ZHANG2024129262} has also been used to enhance the stability of pseudo-labels. However, the lack of target-domain annotations and significant inter-modal gaps challenge accurate pseudo-labels generation and adaptation, posing ongoing challenges for SFDA.

\subsection{Segment Anything Model for Medical Images}
The Segment Anything Model (SAM)~\citep{kirillov2023segment} can deal with segmentation tasks across multiple domains, due to its extensive annotated training set and prompt-based inference mechanism. Its generalizability shows potential for generating pseudo-labels for unlabeled or weakly labeled images~\citep{sunkara2022no}. Recently, SAM has been incorporated into several medical image segmentation studies~\citep{huang2024segment}. For instance, ~\cite{wu2023medical} improved SAM's performance in medical image segmentation through domain-specific fine-tuning with adapters. ~\cite{cheng2024unleashing} explored SAM's application in self-supervised learning with prompt-free fine-tuning for medical image segmentation. 
~\cite{cai2024wspolyp} utilized weakly-supervised learning to adapt SAM for colonoscopy polyp segmentation using pseudo-label generation, while ~\cite{gong20243dsam} employed prompt-based SAM for interactive segmentation of tumors. In addition, the prompt-based inference mechanism has also been applied to segmentation with missing modalities~\citep{DIAO2025128847}.  Despite these advancements, many current SAM-based medical image segmentation methods typically require fine-tuning with a large set of annotated target-domain images~\citep{ma2024segment}. This fine-tuning process is not only time-consuming but also computationally intensive, limiting its practicality for diverse clinical applications. Unlike these works, we explore the application of SAM in the SFDA scenario. Rather than fine-tuning SAM with labeled target-domain images, we leverage SAM's prompt-based inference ability to obtain high-quality pseudo-labels in the target domain, which effectively guides the adaptation of a source model to the unlabeled target domain without access to source-domain images.
\label{sec:meth}
\begin{figure*}
\begin{minipage}[htb]{1.0\linewidth}
  \centering
  \centerline{\includegraphics[width=13.5cm]{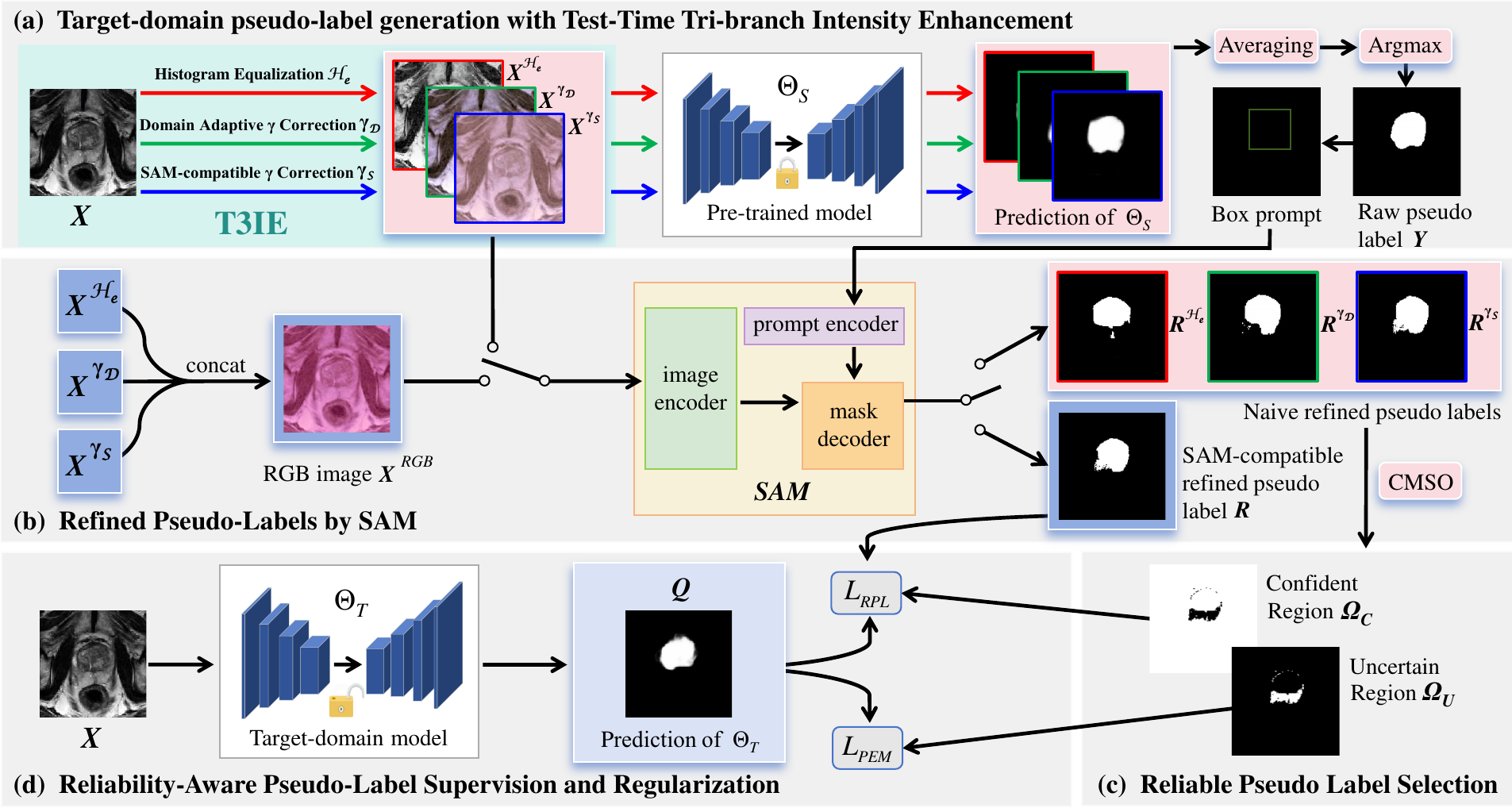}}
\caption{Overview of our proposed SRPL-SFDA method for source-free domain adaptation in medical image segmentation. The arrows in the diagram represent the direction of data flow. Specifically, the red arrows indicate the flow of $X^{\mathcal{H}_e}$, the green arrows represent the flow of $X^{\gamma_D}$, and the blue arrows show the flow of $X^{\gamma_S}$.}
\label{fig1}
\end{minipage}
\end{figure*}

\section{Method}
\label{sec:meth}
Our SRPL-SFDA framework is illustrated in Fig.~\ref{fig1}. Given a model $\Theta_{S}$ pre-trained in a source domain, we aim to adapt it to an unlabeled target domain without access to the source-domain images, and the adapted model is denoted as $\Theta_{T}$. To deal with this problem, we introduce an innovative and efficient iterative pseudo-labeling SFDA method guided by SAM, which consists of four key components: a) Target-domain pseudo-label generation with Test-Time Tri-branch Intensity Enhancement (T3IE), where three different intensity transformations are applied to each target-domain image to mitigate the domain gap and enhance the quality of raw pseudo-labels. b) Pseudo-label refinement by SAM with T3IE-enhanced inputs, where a box prompt is generated from the raw pseudo-label, and the three enhanced images from T3IE are concatenated as a compatible input for SAM to obtain high-quality refined pseudo-labels. c) Reliable pseudo-label mining, where Consistency of Multiple SAM's Outputs (CMSO) under input perturbations with T3IE is proposed to identify reliable and unreliable regions in refined pseudo-labels. d) Pseudo-label learning in the target domain based on an Reliability-Aware Pseudo-Label Supervision and Regularization (RPSR) loss. 

\subsection{Target-Domain Pseudo-Label Generation with T3IE} 
\label{ssec:T3IE}
Given a pre-trained segmentation model $\Theta_{S}$ and an unlabeled target-domain training set, a straightforward approach would be to directly using $\Theta_{S}$ for inference to generate pseudo-labels for the target-domain images. However, due to the distribution shift from source to target domains, this direct inference often results in poor performance. To improve the quality of pseudo-labels, we introduce Test-Time Tri-branch Intensity Enhancement (T3IE) in the target domain, which not only mitigates the domain gap when using the source model to obtain initial pseudo-labels, but also makes them more aligned to the inputs expected by SAM to get refined pseudo-labels. 

\subsubsection{Test-Time Tri-branch Intensity Enhancement (T3IE)}
Specifically, as SAM is designed for RGB images, we consider three adaptive intensity transforms, so that the output of each can serve as a channel for an RGB image. The three intensity transforms are selected to reduce the inter-image variation, and serve as normalization operations in the target domain for robust predictions: histogram equalization, domain adaptive gamma correction, and SAM-compatible gamma correction, and they are described as follows:

\textbf{Histogram Equalization ($\mathcal{H}_e$)}: As medical images often have a low contrast and a skewed distribution of intensities, we apply histogram equalization to redistribute the intensity values of the image, enhancing contrast and making the intensity distribution approximately normalized:
\begin{equation}\label{eq:he}
i' = round\left( (L-1) \cdot CDF_X(i) \right)
\end{equation}
where $CDF_X$ is the cumulative distribution function of image $X$, and $i$ and $i'$ are the input and output intensities for histogram equalization, respectively.  $L$ is the number of gray-scale levels in the image (e.g., 256). 

\textbf{Domain Adaptive Gamma Correction ($\gamma_D$)}: Gamma correction is an effective method to adjust the contrast of an image, but it usually require manual setting of the $\gamma$ value. To automatically determine the $\gamma$ for each image, we align the mean intensity value of an input image to that of the entire dataset, which improves the image contrast and reduces inter-image variation. Let $u_x$ denote the mean intensity of image $X$, and $u_D$ denote that for the entire target-domain dataset, respectively. The target gamma value is denoted as $\gamma_D$, so that $(u_x/L)^{\gamma_D} = u_D/L$, which leads $\gamma_D$ to be:

\begin{equation}\label{1.3}
\gamma_D = \frac{log(u_D/L)}{log(u_X/L)} 
\end{equation}


\textbf{SAM-compatible Gamma Correction ($\gamma_S$)}: As SAM is trained on large-scale natural images, we also adjust the intensity distribution of $X$ to match that of general natural images.  A large corpus of natural images' pixel values are approximately 
distributed with a mean of 127.5 and a standard deviation of 74~\citep{Sener_2022_CVPR}, corresponding to 0.5 and 0.29 after normalization, respectively. These values represent the overall distribution of pixel values across a large dataset of natural images, rather than specific values for individual images. The choice of $\mu$=0.5 and $\sigma$=0.29 reflects the average distribution of natural images, which is stable across the dataset. Although individual images may have varying statistics or distributions, aligning the input intensity distribution to these global values helps in better matching the characteristics of typical natural images, ensuring consistency across the target domain during the adaptation process. To this end, we find an optimal gamma value $\gamma_S$ for gamma correction so that the output's intensity  match this distribution as much as possible:
\begin{equation}
   \gamma_S = \arg\min_{\gamma} \left[ \left(\mu_S - \frac{1}{N} \sum_{i=1}^{N} \hat{X}_i^{\gamma} \right)^2 + \left(\sigma_S^2 - \frac{1}{N} \sum_{i=1}^{N} (\hat{X}_i^{\gamma} - \hat{\mu})^2 \right)^2 \right]
\end{equation}
where $N$ is the number of pixels in image $X$. $\hat{X}_i$ is the normalized intensity value at pixel $i$ in $X$, and $\hat{\mu} = \frac{1}{N}\sum_{i=1}^N\hat{X}_i^\gamma$ is the mean intensity after gamma correction.  $\mu_S$=0.5 and $\sigma_S$=0.29 are the SAM-compatible mean and standard deviation as mentioned above, respectively.

\subsubsection{Initial Pseudo-Labels from Source Model }
For a target-domain image $X$, the outputs after the three intensity transforms of T3IE are denoted as $X^{\mathcal{H}_e}$, $X^{\gamma_D}$ and $X^{\gamma_S}$, respectively. 
These outputs of T3IE are used as inputs for both the source model and SAM to obtain initial pseudo-labels and refined pseudo-labels, respectively. To obtain initial pseudo-labels, we send $X^{\mathcal{H}_e}$, $X^{\gamma_D}$ and $X^{\gamma_S}$ to the source model respectively, and the corresponding probability outputs are denoted as 
$P^{\mathcal{H}_e} = f(\Theta_{S}, X^{\mathcal{H}_e})$, $P^{\gamma_D} = f(\Theta_{S}, X^{\gamma_D})$, and $P^{\gamma_S} = f(\Theta_{S}, X^{\gamma_S})$, respectively. 
The average probability prediction based on T3IE is denoted as $\bar{P}$, and the corresponding initial pseudo-label after $argmax$ operation is denoted as $Y$:
\begin{equation}\label{1.5}
\bar{P} = (P^{\mathcal{H}_e} + P^{\gamma_D}+ P^{\gamma_S})/3;~Y=\text{argmax}{(\bar{P})}
\end{equation}


\subsubsection{Refined Pseudo-Labels by SAM}
\label{ssec:pseudo_label_by_sam}
Though T3IE-based inference with the source model has a better performance than directly using $X$ for inference, the quality of initial pseudo-label $Y$ is still limited. However, $Y$ can roughly provide the location of the segmentation target in target-domain images, and can be used as a prompt of SAM to obtain a refined pseudo-label. 
Specifically, we calculate the bounding box of $Y$ and expand it with a small margin as the bounding box prompt for SAM, which is denoted as $\mathcal{B}$. 



As SAM generally works better on RGB images than gray-scale images~\citep{huang2024segment}, to better leverage SAM to obtain refined pseudo-labels, we convert the input image $X$ into an RGB image by concatenating the three outputs of T3IE:
\begin{equation}\label{1.7}
X^{\text{RGB}} = \text{concat}(X^{\mathcal{H}_e}, X^{\gamma_D}, X^{\gamma_S})
\end{equation}

$X^{\text{RGB}}$ is then sent into SAM to obtain the refined pseudo-label $R$, with $\mathcal{B}$ as the prompt:
\begin{equation}\label{eq:y1}
R = SAM(X^{\text{RGB}}, \mathcal{B})
\end{equation}

Compared with directly sending $X$ or each of the outputs by T3IE into SAM, $R$ obtained by Eq.\eqref{eq:y1} is superior due to two reasons. First, $X^{\text{RGB}}$ is an RGB image that is more compatible with SAM's training image for better performance. Second, the complementary information of different augmented versions of $X$ can be leveraged in the multi-channel-based inference to obtain more robust results. 



\subsection{Reliable Pseudo-Label Mining for Model Adaptation }
\label{ssec:pseudo_label}
Though the refined pseudo-label $R$ has an improved quality by leveraging the knowledge from SAM, it may inherently contain noise due to two factors. 
First, the box prompt is generated from the initial pseudo-label $Y$, where the low quality of $Y$ may limit  accuracy of the prompt, and SAM's good performance relies highly on high-quality prompts. Second, the inference performed by SAM may not consistently delineate object boundaries accurately, especially in  regions with inhomogeneous intensity and complex boundary, due to the low contrast of medical images and the domain gap between natural and medical images. 
Utilizing these noisy labels directly for training can significantly limit the model's performance in the target domain. To address this challenge, we propose Reliability-Aware Pseudo-Label Supervision and R to select reliable pseudo-labels, and then introduce an Reliability-aware Pseudo-label Supervision and Regularization  (RPSR) loss for adaptation in the target domain.

\subsubsection{Consistency of Multiple SAM Outputs (CMSO)}
Inspired by uncertainty estimation via test-time augmentation~\citep{WANG201934} that can obtain more calibrated uncertainty estimation results than a model's raw probability prediction, we leverage the augmented inputs via T3IE to identify reliable regions in the refined pseudo-label $R$.
Specifically, we send $X^{\mathcal{H}_e}$, $X^{\gamma_D}$ and $X^{\gamma_S}$ to SAM respectively with the same bounding box prompt $\mathcal{B}$ derived from $Y$. The corresponding outputs are denoted as $R^{\mathcal{H}_e} = SAM(X^{\mathcal{H}_e}, \mathcal{B})$, $R^{\gamma_D} = SAM(X^{\gamma_D}, \mathcal{B})$, and $R^{\gamma_S} = SAM(X^{\gamma_S}, \mathcal{B})$, respectively. The region with a consensus among the three outputs are treated as reliable part, which is denoted as $\Omega_C$. Correspondingly, the unreliable region is denoted as $\Omega_U$. They are defined as: 
\begin{equation}\label{1.8}
{\Omega _C} = {\{ i  \mid  i \in {\Omega _X}, R^{\mathcal{H}_e}_i = R^{\gamma_D}_i = R^{\gamma_S}_i \}} \end{equation}
\begin{equation}\label{1.9} 
{\Omega _U} = {\Omega _X} \setminus {\Omega _C} \end{equation}
where $i$ is a pixel, and $\Omega_X$ is the set of all pixels in image $X$. 

\subsubsection{Reliability-Aware Pseudo-Label Supervision and Regularization (RPSR)}
\label{ssec:uncertainty}
Building on the refined pseudo-labels and their corresponding reliable regions, we design an reliability-aware pseudo-label supervision loss for training the model in the target domain, and it is combined with a conditional entropy minimization loss for regularization on unreliable regions. For the reliable region $\Omega_C$, we define a reliable pseudo-label loss as:
\begin{equation}\label{1.10}
L_{\text{RPL}}(Q, R) = \frac{1}{2} \left( L_{\text{pce}}(Q, R) + L_{\text{pdc}}(Q, R) \right) \end{equation}
where $Q$ is the probability map predicted by the target-domain model $\Theta_{T}$, and $L_{\text{pce}}$ and $L_{\text{pdc}}$ are the partial cross-entropy and partial Dice losses defined on $\Omega_C$, respectively:
\begin{equation}\label{1.11}
L_{\text{pce}}(Q, R) = - \frac{1}{|\Omega_C|} \sum_{i \in \Omega_C} \sum_{c=0}^{C-1} r_{c,i} \log q_{c,i} \end{equation}
\begin{equation}\label{1.12}
L_{\text{pdc}}(Q, R) = 1 - \frac{1}{C} \sum_{c=0}^{C-1} \frac{2 \sum_{i \in \Omega_C} q_{c,i} r_{c,i} + \varepsilon}{\sum_{i \in \Omega_C} (q_{c,i}^2 + r_{c,i}^2) + \varepsilon} \end{equation}
where $q_{c,i}$ and $r_{c,i}$ are the probability values for class $c$ at pixel $i$ in $Q$ and $R$, respectively, and $\varepsilon$ is a small constant for numerical stability. For the unreliable region $\Omega_U$, we apply a partial entropy minimization loss for regularization:
\begin{equation}\label{1.13}
L_{\text{PEM}}(Q) = - \frac{1}{|\Omega_U|} \sum_{i \in \Omega_U} \sum_{c=0}^{C-1} q_{c,i} \log q_{c,i} \end{equation}

The total loss for adaptation in the target domain is:
\begin{equation}\label{1.14}
L_{\text{total}} = L_{\text{RPL}}(Q, R) + \lambda L_{\text{PEM}}(Q) \end{equation}
where $\lambda$ controls the weight of the partial entropy minimization loss. Note that $\Theta_T$ is initialized with $\Theta_S$.

\section{Experiments and Results}
\subsection{Datasets}
In this study, we evaluated the performance of our proposed SRPL-SFDA on two multi-center datasets: a publicly available Prostate MRI segmentation dataset~\citep{liu2020ms} and an in-house Fetal Brain segmentation dataset~\citep{wu2023upl1}. Table~\ref{table1} provides a detailed summary of the key characteristics of these datasets, including imaging modalities and setting for source and target domains.

\subsubsection{Prostate MRI Segmentation Dataset}
The Prostate MRI Segmentation Dataset comprises data from six different institutions (sites A to F) for prostate segmentation, and data from site C was excluded as it contains data from unhealthy patients. The dataset includes a total of 97 3D volumes, with 30, 30, 13, 12 and 12 volumes from site A,  B, D, E and F, respectively. The images from different sites were acquired with field strengths ranging from 1.5 to 3 Tesla, with in-plane resolutions between 0.25 mm and 0.625 mm, slice thicknesses varying from 1.25 mm to 4 mm, and through-plane resolutions from 2.2 mm to 4.0 mm. Additionally, varying coil types were used, including surface and endorectal coils, with the manufacturers represented by Siemens, Philips, and GE across the different institutions.
We combined data from sites A and B to form the source domain, while data from sites D, E and F were designated as the target domain.  The images from each domain were randomly split into 70\% for training, 10\% for validation, and 20\% for testing, with labels for the training set in the target domain intentionally excluded to simulate an unsupervised domain adaptation scenario.

\subsubsection{Fetal Brain Segmentation Dataset}
The Fetal Brain dataset~\citep{wu2023upl1} consists of fetal MRI scans acquired at a single medical center using two distinct imaging protocols. It includes 44 volumes obtained with True Fast Imaging with Steady-State Precession (TrueFISP) and 68 volumes acquired using Half-Fourier Acquisition Single-shot Turbo spin Echo (HASTE). The number of slices per volume varies between 11 and 22, with the gestational ages of the fetuses ranging from 21 to 33 weeks. The in-plane resolution for TrueFISP spans from 0.67 to 1.12 mm, and from 0.64 and 0.70 mm for the HASTE sequence. The slice thickness varies between 6.5 and 7.15 mm. In our experiments, the TrueFISP and HASTE sequences were used as source and target domains, respectively.
We randomly partitioned the images within each domain into training, validation, and testing sets, at a ratio of 7:1:2 at the patient level. Labels for the training images in the target domain (HASTE) were intentionally excluded in the experiment.

\begin{table}
  \centering
  \caption{Details of the experimental datasets. The reported values correspond to the number of volumes within each dataset.}
  \scalebox{0.91}{
    \begin{tabular}{c|ccc|cc}
    \toprule
    Dataset & \multicolumn{3}{c|}{Prostate MRI dataset} & \multicolumn{2}{c}{Fetal brain dataset} \\
    \midrule
    \multirow{2}[4]{*}{Domain} & Source & Target 1 & Target 2  & Source & Target \\
\cmidrule{2-6}          &  Site A, B  &  Site D, E, F & Site C  & TrueFISP & HASTE \\
    \midrule
    Training & 42    & 25  & 13   & 30    & 47 \\
    Validation & 6     & 4   & 2    & 5     & 7 \\
    Testing & 12     & 8    & 4   & 9     & 14 \\
    \midrule
    Overall & 60    & 37  & 19   & 44    & 68 \\
    \bottomrule
    \end{tabular}  }%
  \label{table1}%
\end{table}%
\subsection{Implementation Details and Evaluation Metrics}
In our experiments, we first normalized the image intensities of both datasets to the range [0, 1]. Additionally, all images were resampled to a resolution of 1×1×1$mm^{3}$. Given the variation in slice thickness across the datasets, we employed U-Net~\citep{ronneberger2015u} as the backbone segmentation network to perform slice-level segmentation,  with the results of slices  stacked into 3D volumes for final evaluation. The input patch size for both datasets was set to 256 × 256. Our method was implemented using the PyMIC\footnote{https://github.com/HiLab-git/PyMIC}~\citep{Wang2022pymic} library and PyTorch, with all experiments conducted on an NVIDIA RTX 3090 GPU.

To fully evaluate SRPL-SFDA’s segmentation performance, we used two common metrics, Dice coefficient and Average Symmetric Surface Distance (ASSD). They are defined as:
\begin{equation}\label{1.15}
\text{DICE} = \frac{2|A \cap A'|}{|A| + |A'|} \end{equation}
where A and A' represent the predicted and ground truth segmentation volumes, respectively.
\begin{equation}\label{1.16}
\text{ASSD} = \frac{1}{|S| + |G|} \left( \sum_{s \in S} d(s, G) + \sum_{g \in G} d(g, S) \right)
 \end{equation}
where $S$ and $G$ represent the sets of surface points from the segmentation result and the ground truth, respectively. $d(s,G)$ denotes the shortest Euclidean distance between a point $s\in S$ and any point in $G$.

For training the source-domain model, we utilized a combination of cross-entropy and Dice loss functions, and employed the Adam optimizer with a batch size of 64, an initial learning rate of $10^{-3}$ and 400 epochs. The learning rate was reduced to 1\% every 40 epochs. The checkpoint with the best performance on the source-domain validation set was selected to initialize the target-domain model.

During the adaptation phase, we used the Adam optimizer with a fixed learning rate of $6.0 \times {10^{ - 4}}$, and an epoch number of 300. The batch size was set to 64. According to the best performance on the validation set of the target domain, the hyper-parameter setting was $\lambda$ = 10.0, and the corresponding checkpoint was employed for inference on  testing images of each target domain. The segmentation performance was quantitatively evaluated using the Dice coefficient and Average Symmetric Surface Distance (ASSD) at volume level between a segmentation output and the ground truth.
\label{sec:meth}
\begin{figure*}
\begin{minipage}[htb]{1\linewidth}
  \centering
  \centerline{\includegraphics[width=15cm]{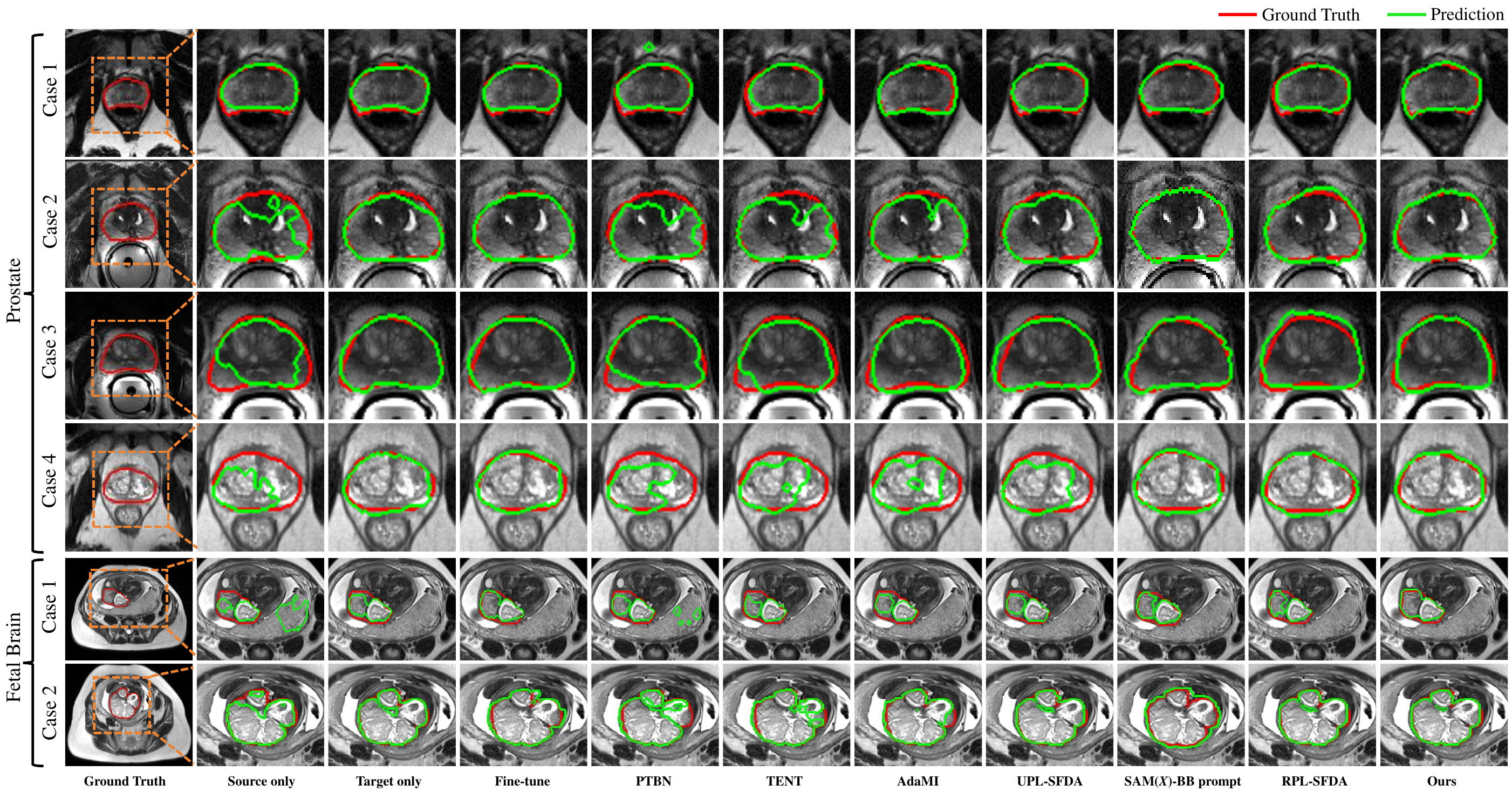}}
\caption{Visual comparison between different SFDA methods on the prostate and fetal brain segmentation tasks.}
\label{fig3}
\end{minipage}
\end{figure*}
\subsection{Comparison with State-of-the-art SFDA Methods}
Our method was evaluated against four state-of-the-art SFDA methods: 1) \textbf{PTBN}~\citep{nado2020evaluating} that adapts batch normalization statistics of unlabeled target-domain images images without using loss functions for optimization; 2) \textbf{TENT}~\citep{wang2020tent} that updates batch normalization parameters by minimizing the entropy of model predictions in the target domain; 3) \textbf{AdaMI}~\citep{bateson2022source} that maximizes adaptive mutual information between the source and target domains; 
and 4) \textbf{UPL-SFDA}~\citep{wu2023upl} that employs target-domain growing of decoders to obtain pseudo-labels for guiding the model adaptation. 

In addition, we compared our method with its three variants: 1) \textbf{ SAM($X$)-BB prompt}, which generates a bounding box by randomly expanding 5-10 pixels from the ground truth of the target domain to simulate user interaction as a prompt, and combines this with the original image $X$ for input into SAM for segmentation; 2)  the preliminary version \textbf{RPL-SFDA}~\citep{liu2024rpl} that obtains pseudo-labels based on random non-linear intensity augmentations without using SAM; 3) \textbf{SRPL-SFDA (scratch)} that 
initializes $\Theta_T$ from scratch and uses the pseudo-labels obtained by $\Theta_S$ and refined by SAM for  training in the target domain. Lastly, we also experimented with three supervised methods for reference: 1) “\textbf{Source only}” that directly applies the source model for inference in the target domain; 2) “\textbf{Target only}” that trains the model solely with annotated target-domain images; and 3) “\textbf{Fine-tune}” that involves fine-tuning the source model using annotated target-domain training images. To ensure a fair comparison, all the compared methods were implemented with the same source model. 

\subsubsection{Results on Prostate MRI Segmentation Dataset}
Table~\ref{table2} shows the quantitative comparison among different methods on the prostate MRI segmentation dataset. Firstly, for the supervised segmentation methods, it is evident that  ``Target only" substantially outperformed  ``Source only", highlighting the significant domain gap between the source and target domains. Specifically, ``Source only" achieved an average Dice of 69.88\% in the target domain, while ``Target only" obtained an average Dice of 83.02\%. 

\begin{table}
  \centering
  \caption{Quantitative evaluation of different SFDA methods on the Prostate MRI Dataset. $^{*}$ indicates a  significant improvement over the best existing SFDA method using a paired Student's t-test ($p$-value $<$ 0.05).}
  \scalebox{0.7}{
    \begin{tabular}{cc|cccccc|cccccc}
    \toprule
    \multicolumn{2}{c|}{\multirow{2}[2]{*}{Method}}  & \multicolumn{6}{c|}{Site D, E, F}          & \multicolumn{6}{c}{Site C}  \\
    \cmidrule{3-14}   \multicolumn{2}{c|}{} & \multicolumn{3}{c}{Dice (\%)} & \multicolumn{3}{c|}{ASSD (mm)} & \multicolumn{3}{c}{Dice (\%)} & \multicolumn{3}{c}{ASSD (mm)} \\
    \midrule
    \multicolumn{2}{c|}{Source only} & \multicolumn{3}{c}{69.88 ± 19.35} & \multicolumn{3}{c|}{5.20 ± 3.41} & \multicolumn{3}{c}{58.98 ± 19.09} & \multicolumn{3}{c}{5.61 ± 3.56}\\
    \multicolumn{2}{c|}{Target only} & \multicolumn{3}{c}{83.02 ± 6.96} & \multicolumn{3}{c|}{3.08 ± 1.54}  & \multicolumn{3}{c}{80.97 ± 7.87} & \multicolumn{3}{c}{2.19 ± 0.84} \\
    \multicolumn{2}{c|}{Fine-tune} & \multicolumn{3}{c}{85.30 ± 5.03} & \multicolumn{3}{c|}{2.50 ± 1.51} & \multicolumn{3}{c}{81.60 ± 4.15} & \multicolumn{3}{c}{1.64 ± 0.21} \\
    \midrule
    \multicolumn{2}{c|}{PTBN\hspace{1.5mm}~\citep{nado2020evaluating}} & \multicolumn{3}{c}{71.97 ± 19.16} & \multicolumn{3}{c|}{4.53 ± 2.87} & \multicolumn{3}{c}{63.87 ± 5.21} & \multicolumn{3}{c}{3.07 ± 2.08}\\
    \multicolumn{2}{c|}{TENT\hspace{1.5mm}~\citep{wang2020tent}} & \multicolumn{3}{c}{72.18 ± 18.99} & \multicolumn{3}{c|}{4.59 ± 2.85}& \multicolumn{3}{c}{67.70 ± 6.80} & \multicolumn{3}{c}{2.83 ± 1.98} \\
    \multicolumn{2}{c|}{AdaMI\hspace{1.5mm}~\citep{bateson2022source}} & \multicolumn{3}{c}{76.85 ± 11.73} & \multicolumn{3}{c|}{4.84 ± 2.31} & \multicolumn{3}{c}{68.20 ± 4.79} & \multicolumn{3}{c}{2.60 ± 1.77} \\
    \multicolumn{2}{c|}{UPL-SFDA\hspace{1.5mm}~\citep{wu2023upl}} & \multicolumn{3}{c}{77.27 ± 7.38} & \multicolumn{3}{c|}{4.83 ± 2.90}& \multicolumn{3}{c}{73.55 ± 2.76} & \multicolumn{3}{c}{2.46 ± 0.70} \\
    \multicolumn{2}{c|}{SAM($X$)-BB prompt} & \multicolumn{3}{c}{80.39 ± 9.17} & \multicolumn{3}{c|}{2.45 ± 0.66}& \multicolumn{3}{c}{77.03 ± 2.55}& \multicolumn{3}{c}{2.21 ± 0.64} \\
    \multicolumn{2}{c|}{RPL-SFDA\hspace{1.5mm}~\citep{liu2024rpl}} & \multicolumn{3}{c}{81.02 ± 6.08} & \multicolumn{3}{c|}{2.31 ± 0.76}& \multicolumn{3}{c}{78.36 ± 2.74} & \multicolumn{3}{c}{2.20 ± 0.76} \\
    \multicolumn{2}{c|}{SRPL-SFDA (scratch)} & \multicolumn{3}{c}{78.29 ± 5.83} & \multicolumn{3}{c|}{2.81 ± 1.39}& \multicolumn{3}{c}{76.15 ± 2.29} & \multicolumn{3}{c}{2.23 ± 0.62} \\
    \multicolumn{2}{c|}{SRPL-SFDA} & \multicolumn{3}{c}{\textbf{82.22 ± 3.94$^{*}$}} & \multicolumn{3}{c|}{\textbf{2.02 ± 0.50}}& \multicolumn{3}{c}{\textbf{80.12 ± 1.52$^{*}$}} & \multicolumn{3}{c}{\textbf{1.86 ± 0.41}}  \\
    \bottomrule
    \end{tabular} }%
  \label{table2}%
\end{table}%

In terms of existing SFDA methods, PTBN~\citep{nado2020evaluating} and TENT~\citep{wang2020tent} obtained a slight improvement from ``Source only", with an average Dice of 71.97\% and 72.18\%, respectively. AdaMI~\citep{bateson2022source} and UPL-SFDA~\citep{wu2023upl} exhibited a noticeable improvement, with average Dice being 76.85\% and 77.27\%, respectively.  In contrast, our  method demonstrated significant superiority over the existing SFDA methods, with an average Dice of 82.22\%, and ASSD of 2.02~mm, respectively. Furthermore, compared to the ``Source only" approach, the SAM($X$)-BB prompt results in a 10.51\% improvement in average Dice scores, highlighting SAM's robust generalization ability, minimal sensitivity to domain shifts, and outstanding zero-shot inference performance in the target domain. Note that our preliminary work RPL-SFDA~\citep{liu2024rpl} has already outperformed the other existing SFDA methods. Compared with RPL-SFDA, our method SRPL-SFDA improved the average Dice by 1.2 percentage points, showing the effectiveness of our method using SAM-based pseudo-labels. Compared with SAM($X$)-BB prompt, our method also improves the average Dice by 1.83 percentage points, showing its advantage over directly using SAM for user interaction-based inference. In addition, we found that the performance of our method decreased when training from scratch in the target domain, demonstrating the usefulness of source model for initialization. 

In terms of average ASSD, our method achieved the lowest value of 2.02~mm among the compared SFDA methods, which is significantly better than both ``Fine-tune" (2.50 mm) and ``Target only" (3.08 mm). This reduction in surface distance indicates more accurate boundary delineation, which is critical for clinical applications where accurate boundary is desired. Fig.~\ref{fig3} shows a visual comparison between these methods. In the first case, 
the source model has a relatively good performance due to the good contrast and homogeneous target region. The existing SFDA methods performed similarly to the source model, while our method obtains a boundary that is aligned tightly with the ground truth. In the second case, the ``Source only" method has a large under-segmentation. Existing methods like PTBN and TENT also under-segmented the prostate, while our method obtains a more accurate result than the existing SFDA methods. These results emphasize the effectiveness and robustness of our method, particularly in challenging cases.

Meanwhiles, our approach focuses on improving the adaptability of large models like SAM for segmentation tasks across both natural and medical images. In the Prostate MRI segmentation dataset~\citep{liu2020ms}, site C, derived from cancer patients, was excluded due to its potential to introduce significant biases. Most images (17 out of 19) in site C are acquired from patients with prostate cancer, leading to a semantic difference in the prostate area. These differences above lead to a visible appearance difference among multi-site data, and also lead to an intensity distribution shift~\citep{liu2020ms}. This follows previous work~\cite{tomar2022opttta}, where excluding datasets with unhealthy patient data was shown to reduce unnecessary interference and help highlight the model's performance more clearly. As shown in Table~\ref{table2}, using site C as a separate target domain reveals a noticeable performance decline with the source-only model compared to the target-only model, demonstrating that our method can achieve better performance. Future work could explore bridging not only the gap between SAM’s segmentation of natural and medical images but also the differences between source domains with healthy patients and target domains with unhealthy patients.
\begin{table}
  \centering
  \caption{Quantitative comparison of different SFDA methods on the fetal brain segmentation dataset. $^{*}$ indicates a  significant improvement  over the best existing SFDA method using a paired Student's t-test ($p$-value $<$ 0.05).}
  \scalebox{0.85}{
    \begin{tabular}{cc|cccccc}
    \toprule
    \multicolumn{2}{c|}{Method} & \multicolumn{3}{c}{Dice (\%)} & \multicolumn{3}{c}{ASSD (mm)} \\
    \midrule
    \multicolumn{2}{c|}{Source only} & \multicolumn{3}{c}{83.47 ± 19.61} & \multicolumn{3}{c}{3.80 ± 14.17} \\
    \multicolumn{2}{c|}{Target only} & \multicolumn{3}{c}{95.53 ± 2.09} & \multicolumn{3}{c}{0.52 ± 0.46} \\
    \multicolumn{2}{c|}{Fine-tune} & \multicolumn{3}{c}{96.12 ± 1.15} & \multicolumn{3}{c}{0.51 ± 0.43} \\
    \midrule
    \multicolumn{2}{c|}{PTBN\hspace{1.5mm}~\citep{nado2020evaluating}} & \multicolumn{3}{c}{88.40 ± 8.58} & \multicolumn{3}{c}{4.25 ± 4.50} \\
    \multicolumn{2}{c|}{TENT\hspace{1.5mm}~\citep{wang2020tent}} & \multicolumn{3}{c}{89.48 ± 6.14} & \multicolumn{3}{c}{5.65 ± 9.54} \\
    \multicolumn{2}{c|}{AdaMI\hspace{1.5mm}~\citep{bateson2022source}} & \multicolumn{3}{c}{92.29 ± 3.47} & \multicolumn{3}{c}{5.17 ± 7.33} \\
    \multicolumn{2}{c|}{UPL-SFDA\hspace{1.5mm}~\citep{wu2023upl}} & \multicolumn{3}{c}{92.64 ± 3.55} & \multicolumn{3}{c}{3.94 ± 4.81} \\
    \multicolumn{2}{c|}{SAM($X$)-BB prompt} & \multicolumn{3}{c}{92.78 ± 3.86} & \multicolumn{3}{c}{1.66 ± 3.26} \\
    \multicolumn{2}{c|}{RPL-SFDA\hspace{1.5mm}~\citep{liu2024rpl}} & \multicolumn{3}{c}{93.81 ± 1.71} & \multicolumn{3}{c}{0.87 ± 0.61} \\
    \multicolumn{2}{c|}{SRPL-SFDA (scratch)} & \multicolumn{3}{c}{93.60 ± 2.72} & \multicolumn{3}{c}{1.40 ± 1.78} \\
    \multicolumn{2}{c|}{SRPL-SFDA} & \multicolumn{3}{c}{\textbf{94.33 ± 1.82$^{*}$}} & \multicolumn{3}{c}{\textbf{0.78 ± 0.77}} \\
    \bottomrule
    \end{tabular}   }%
  \label{table3}%
\end{table}%


\subsubsection{Results on Fetal Brain Segmentation Dataset}
Quantitative comparison of different methods on the fetal brain segmentation dataset is listed in Table~\ref{table3}. The average Dice of ``Source only" and ``Target only" was  83.47\% and 95.53\% respectively, underscoring the large gap between source and target domains. The average Dice obtained by  PTBN~\citep{nado2020evaluating} and TENT~\citep{wang2020tent} was 88.40\% and 89.48\%, respectively, while the other SFDA methods obtained Dice scores above 90\%. Specifically, UPL-SFDA~\citep{wu2023upl} and RPL-SFDA~\citep{liu2024rpl} increased the average Dice to 92.64\% and 93.81\%, respectively. In contrast, our method SRPL-SFDA outperformed all the existing SFDA methods, with an average Dice of 94.33\% and ASSD of 0.78~mm. SAM($X$)-BB prompt obtained an average Dice of 92.78\%, showing the strong generalization ability of SAM with user interactions, however, it is still inferior to our method. Note that even SRPL-SFDA (scratch) achieved an average Dice of 93.60\%, which is only inferior to RPL-SFDA~\citep{liu2024rpl} among the existing methods. 
A visual comparison of these methods is shown in the last two rows of Fig.~\ref{fig3}, and it can be observed that segmentation results of our method are closer to the ground truths than the other SFDA methods, and they are also comparable to those obtained by ``Target only".

\subsection{Ablation study}
In this section, we conducted comprehensive ablation studies on the validation set in the target domains for analyzing the effectiveness of our method for pseudo-label refinement and target-domain model training. We also analyzed the hyper-parameter involved in our method. 

\subsubsection{Effectiveness of T3IE and SAM on Pseudo-Label Quality}
To validate the effectiveness of our method in obtaining high-quality pseudo-labels, we compared different variants as shown in Table~\ref{table6}, where $\Theta_S$ means inference with the source model, T3IE means using T3IE for the inputs of $\Theta_S$. SAM($X$) and SAM($X^{RGB}$) mean using the original testing image and the three-channel image after T3IE for inputs of SAM, respectively. Since SAM is designed for natural images, which are typically three-channel, the original single-channel medical image $X$ is duplicated by three times to obtain a three-channel image before being sent into SAM. Additionally, we also conducted experiments where SAM($X$) only used the bounding box obtained by $\Theta_S$ for prompt-based inference, which helps illustrate the difference between the bounding boxes generated from the original input to the source model and those produced using the ensemble method.
\begin{table*}
  \centering
  \caption{Effect of different components of our method on pseudo-label quality in the target domain. SAM($X$) means repeating gray-scale medical image by three times to obtain a three-channel input for SAM.}
  \scalebox{0.75}{
    \begin{tabular}{cccc|cccccc|cccccc}
    \toprule
    \multicolumn{4}{c|}{Components}       &  \multicolumn{6}{c|}{Prostate}          & \multicolumn{6}{c}{Fetal brain} \\
    \midrule
    $\Theta_S$     & \textcolor{black}{T3IE}   & SAM($X$)     & SAM($X^{RGB}$)   & \multicolumn{3}{c}{Dice (\%)} & \multicolumn{3}{c|}{ASSD (mm)} & \multicolumn{3}{c}{Dice (\%)} & \multicolumn{3}{c}{ASSD (mm)} \\
    \midrule
    \checkmark  &       &       &       &  \multicolumn{3}{c}{73.79 ± 12.18} & \multicolumn{3}{c|}{3.68 ± 0.85} & \multicolumn{3}{c}{83.88 ± 13.26} & \multicolumn{3}{c}{3.46 ± 6.92} \\
    \checkmark  & \checkmark      &       &       &  \multicolumn{3}{c}{76.37 ± 10.34} & \multicolumn{3}{c|}{3.50 ± 0.61} & \multicolumn{3}{c}{88.24 ± 10.74} & \multicolumn{3}{c}{1.70 ± 3.44} \\
    \checkmark  &       &    \checkmark   &       &  \multicolumn{3}{c}{78.32 ± 9.81} & \multicolumn{3}{c|}{2.94 ± 1.48} & \multicolumn{3}{c}{90.74 ± 5.06} & \multicolumn{3}{c}{1.69 ± 3.02} \\
      \checkmark  &   \checkmark    &  \checkmark     &       &  \multicolumn{3}{c}{79.54 ± 9.75} & \multicolumn{3}{c|}{2.60 ± 1.63} & \multicolumn{3}{c}{91.51 ± 4.82} & \multicolumn{3}{c}{1.66 ± 3.19} \\
      \checkmark  &   \checkmark    &      &  \checkmark     &  \multicolumn{3}{c}{\textbf{81.46 ± 9.46}} & \multicolumn{3}{c|}{\textbf{2.49 ± 0.71}} & \multicolumn{3}{c}{\textbf{91.74 ± 5.07}} & \multicolumn{3}{c}{\textbf{1.45 ± 2.06}} \\
    \bottomrule
    \end{tabular}  }%
  \label{table6}%
\end{table*}%
\label{sec:meth}

Table~\ref{table6} demonstrates that each component of our method helped to improve the quality of pseudo-labels.  Taking the prostate  dataset as an example, only using $\Theta_S$ for inference yielded a Dice score of 73.79\%, and introducing T3IE improved it to 76.37\%. Using the bounding box obtained by $\Theta_S$ as prompt of SAM on the original image for refinement improved it to 78.32\%. Using the corresponding bounding box obtained by $\theta_S$ + T3IE as prompt of SAM on the original image improved it to 79.54\%. Finally, replacing the input image with the T3IE-augmented version for SAM further improved it to 81.46\%, which is an improvement of 7.67 percentage points compared with the source model. On the fetal brain dataset, Table~\ref{table6} also shows that each component of our method helped to improve the average Dice and reduce the ASSD of pseudo-labels. Compared with the source model, our method obtained a gain of 7.86 percentage points in terms of average Dice.
The results  demonstrate that using SAM in bounding box mode, coupled with augmented inputs with T3IE, significantly enhances the accuracy of pseudo-labels, and highlight the benefits of using SAM-compatible inputs for pseudo-label refinement. 
\begin{figure*}
\begin{minipage}[htb]{1.0\linewidth}
  \centering
  \centerline{\includegraphics[width=11cm]{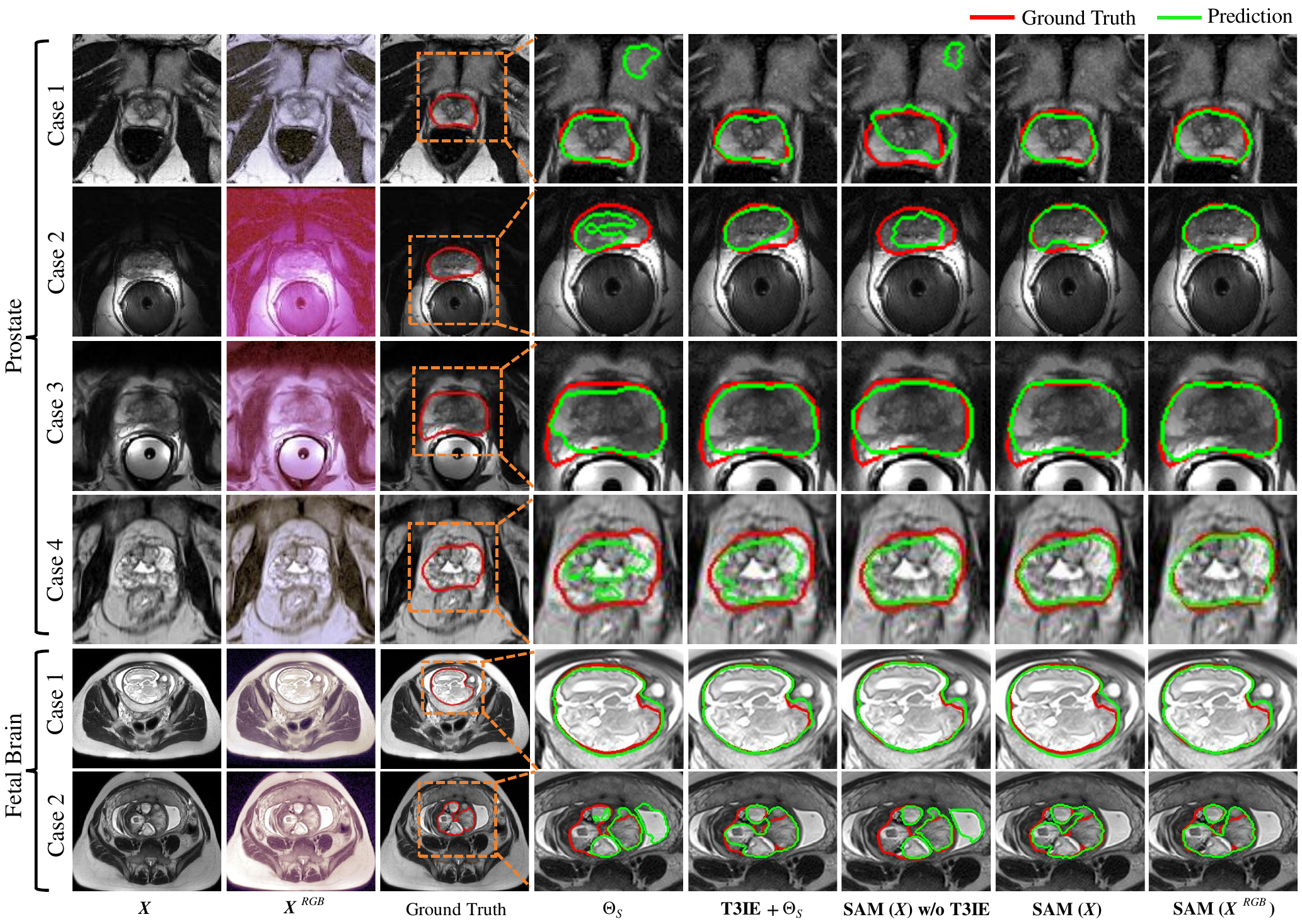}}
\caption{Visual comparison of pseudo-labels in the target domain obtained by  different methods.}
\label{fig5}
\end{minipage}
\end{figure*}
Fig.~\ref{fig5} shows a visual comparison of pseudo-labels obtained by different methods. It  demonstrates that inference with $\Theta_S$ obtained the lowest quality, and $\Theta_S$ with T3IE largely improved the accuracy of  pseudo-labels. The results obtained by SAM were better than those obtained by $\Theta_S$, and compared with SAM($X$), SAM($X^{RGB}$) led to more accurate boundaries in the pseudo-labels. 

\subsubsection{Effectiveness of CMSO and RPSR in Pseudo-Label Learning}
We also conducted an ablation study with different variants of our pseudo-label learning method: 1) \textbf{``EM"} represents the use of entropy minimization di
rectly on the predictions without leveraging pseudo-labels; 2) \textbf{``PL($Y$)"} denotes using initial pseudo-label $Y$ obtained by the source-model with T3IE for fully supervised learning; 3) \textbf{``PL($R$)"} denotes using the refined pseudo-label $R$ based on SAM with $X^{RGB}$ for fully supervised learning; 4) \textbf{``RPL"} means applying pseudo-labels supervision only to reliable regions identified by CMSO; 
and 5) \textbf{``RPL + PEM"} corresponds to the loss defined in Eq.~\eqref{1.14} of our method. 

\begin{table*}
  \centering
  \caption{Ablation study of our method when training with pseudo-labels in the target domain.}
  \scalebox{0.9}{
    \begin{tabular}{cc|cccccc|cccccc}
    \toprule
    \multicolumn{2}{c|}{\multirow{2}[2]{*}{Method}}  & \multicolumn{6}{c|}{Prostate}          & \multicolumn{6}{c}{Fetal brain } \\
    \cmidrule{3-14}   \multicolumn{2}{c|}{} & \multicolumn{3}{c}{Dice (\%)} & \multicolumn{3}{c|}{ASSD (mm)} & \multicolumn{3}{c}{Dice (\%)} & \multicolumn{3}{c}{ASSD (mm)} \\
    \midrule
    \multicolumn{2}{c|}{EM} & \multicolumn{3}{c}{74.15 ± 13.33} & \multicolumn{3}{c|}{6.27 ± 3.25} & \multicolumn{3}{c}{87.86 ± 13.29 } & \multicolumn{3}{c}{6.74 ± 6.93} \\
    \multicolumn{2}{c|}{PL($Y$)} & \multicolumn{3}{c}{80.67 ± 5.07} & \multicolumn{3}{c|}{3.91 ± 0.89} & \multicolumn{3}{c}{88.35 ± 6.03} & \multicolumn{3}{c}{2.61 ± 2.90} \\
    \multicolumn{2}{c|}{PL($R$)} & \multicolumn{3}{c}{82.49 ± 4.81} & \multicolumn{3}{c|}{3.37 ± 2.74} & \multicolumn{3}{c}{88.72 ± 5.95} & \multicolumn{3}{c}{2.35 ± 1.48} \\
    \multicolumn{2}{c|}{RPL} & \multicolumn{3}{c}{83.27 ± 4.99} & \multicolumn{3}{c|}{3.54 ± 2.14} & \multicolumn{3}{c}{91.84 ± 5.28} & \multicolumn{3}{c}{1.54 ± 1.37} \\
    \multicolumn{2}{c|}{Ours (RPL + PEM)} & \multicolumn{3}{c}{\textbf{84.46 ± 5.23}} & \multicolumn{3}{c|}{\textbf{2.45 ± 0.68}} & \multicolumn{3}{c}{\textbf{92.58 ± 4.27}} & \multicolumn{3}{c}{\textbf{1.13 ± 1.06}} \\
    \bottomrule
    \end{tabular}    }%
  \label{table5}%
\end{table*}%
Table~\ref{table5} demonstrates the performance improvement associated with each component. ``EM" obtained the lowest performance, with average Dice score of 74.15\% and 87.86\% on the prostate and fetal brain segmentation datasets, respectively. 
Introducing pseudo-labels with ``PL($Y$)" improves the Dice score  to 80.67\% for the prostate and 88.35\% for the fetal brain, respectively. 
``PL($R$)" further improves the corresponding Dice values to 82.49\% and 88.72\%, respectively, showing the superiority of refined pseudo-labels obtained by SAM.  
In addition, RPL outperformed ``PL($R$)", demonstrating the superiority of supervised learning with the reliable region of pseudo-labels. Finally, combining RPL and PEM led to the best performance, with an average Dice of 84.46\% and 92.58\% on the two datasets, respectively, further showing the effectiveness of entropy minimization regularization on the unreliable region of pseudo-labels. The results in Table~\ref{table5} also show that each component of our loss function helped to reduce the ASSD values with more accurate boundaries in the segmentation outputs.
\label{sec:meth}
\begin{figure}
\begin{minipage}[htb]{1.0\linewidth}
  \centering
  \centerline{\includegraphics[width=8cm]{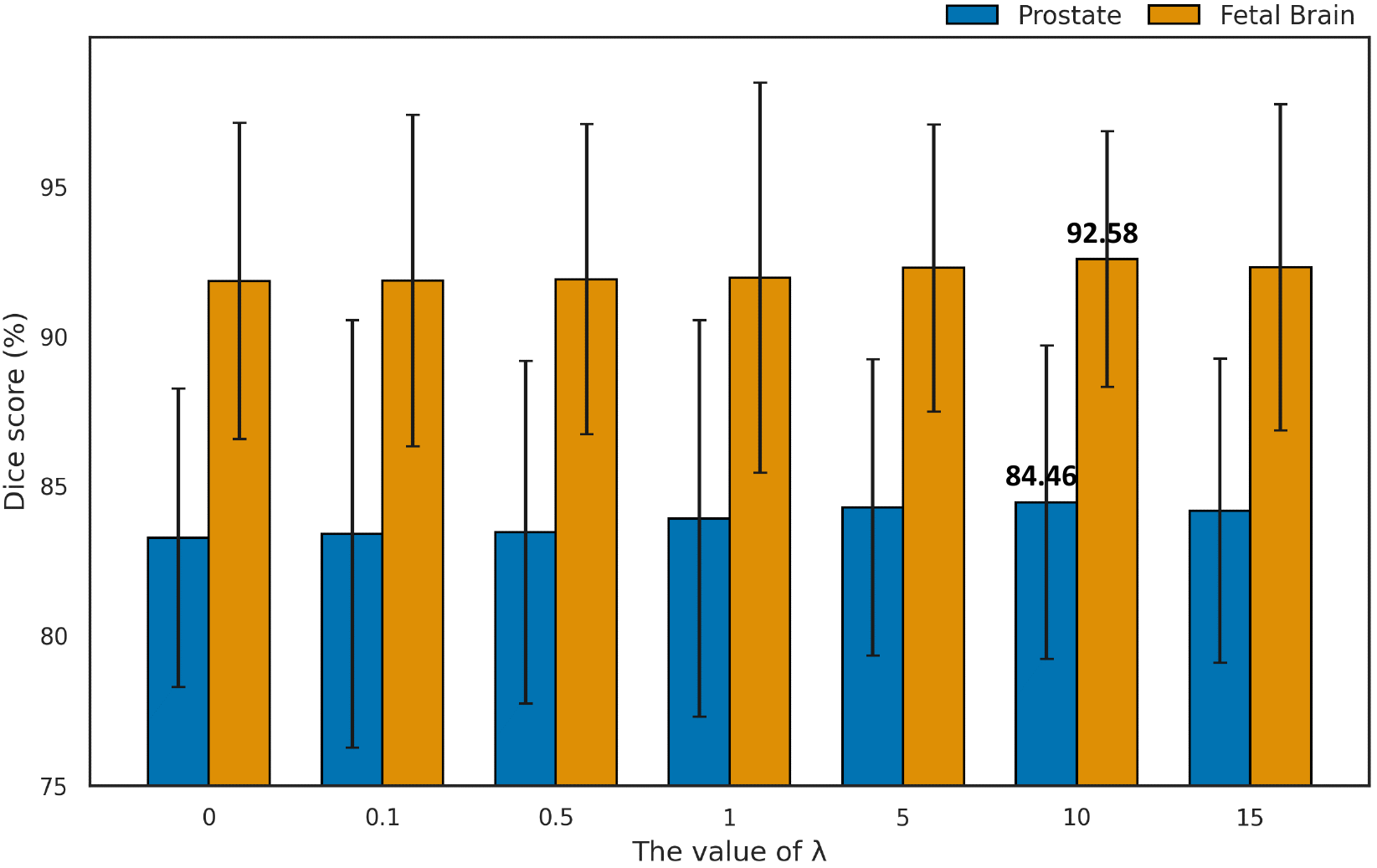}}
\caption{Effect of different $\lambda$ values  on the validation sets of different target domains.}
\label{fig6}
\end{minipage}
\end{figure}
\subsubsection{Hyper-parameter Analysis}
Our method only has one hyper-parameter $\lambda$ that balances the weight of $L_{PEM}$ in Eq.~\eqref{1.14}. Fig.~\ref{fig6} shows the results on the validation dataset with different $\lambda$ values. Note that $\lambda=0$ corresponds to only using $L_{RPL}$ loss in Eq.~\eqref{1.14}. Fig.~\ref{fig6} shows that setting $\lambda>0$ generally performes better than $\lambda=0$, demonstrating the effectiveness of $L_{PEM}$ on unreliable pseudo-label regions. However, a $\lambda$ that is too large will weaken the supervision effect from $L_{RPL}$. According to Fig.~\ref{fig6}, $\lambda=10.0$ obtained the best performance on both datasets, and therefore we set $\lambda=10.0$ in the experiments.

\subsubsection{Effect of Different Prompt Types for SAM}
We conducThese values represent the overall distributioe performance of SAM under different prompt types. Specifically, we evaluated three configurations: \textbf{1) SAM($X$)-mask}, where the pseudo-label from the source model is directly used as a prompt; \textbf{2) SAM($X$)-point}, where 5-10 random points are generated within the pseudo-label region as prompts; and \textbf{3) SAM($X$)-box}, where a bounding box of the pseudo-label expanded by a small margin (i.e., 5-10 pixels) is used as the prompt.

The results, as shown in Table~\ref{table8}, demonstrate that SAM($X$)-box consistently outperforms the other two configurations across both datasets in terms of Dice and ASSD metrics. For the Prostate dataset, SAM($X$)-box achieves a Dice score of 78.32\% and an ASSD of 2.94 mm, significantly better than those of SAM($X$)-point and SAM($X$)-mask. A similar trend is observed on the Fetal brain dataset, where SAM($X$)-box achieves a Dice score of 90.74\% and an ASSD of 1.69 mm, compared to 81.62\% and 10.37 mm for SAM($X$)-point, and 85.16\% and 1.70 mm for SAM($X$)-mask.

The superior performance of SAM($X$)-box can be attributed to two key factors. First, bounding box prompts are more robust compared to fine-grained prompts like masks or points, as they are less sensitive to noise and inaccuracies in the pseudo-labels. Second, the use of bounding boxes minimizes the impact of low-quality pseudo-labels, as the prompts are based on the general localization of the target region rather than its precise contour. This makes bounding box prompts more suitable for scenarios where pseudo-label quality is inherently limited, as is often the case in domain adaptation settings.
\begin{table*}
  \centering
  \caption{Ablation study on prompt types for SAM when training with pseudo-labels in the target domain.}
  \scalebox{0.9}{
    \begin{tabular}{cc|cccccc|cccccc}
    \toprule
    \multicolumn{2}{c|}{\multirow{2}[2]{*}{Method}}  & \multicolumn{6}{c|}{Prostate}          & \multicolumn{6}{c}{Fetal brain } \\
    \cmidrule{3-14}   \multicolumn{2}{c|}{} & \multicolumn{3}{c}{Dice (\%)} & \multicolumn{3}{c|}{ASSD (mm)} & \multicolumn{3}{c}{Dice (\%)} & \multicolumn{3}{c}{ASSD (mm)} \\
    \midrule
    \multicolumn{2}{c|}{SAM($X$)-mask} & \multicolumn{3}{c}{56.51 ± 9.86} & \multicolumn{3}{c|}{10.04 ± 5.68} & \multicolumn{3}{c}{85.16 ± 5.99} & \multicolumn{3}{c}{1.70 ± 3.88} \\
    \multicolumn{2}{c|}{SAM($X$)-point} & \multicolumn{3}{c}{68.27 ± 10.34} & \multicolumn{3}{c|}{11.37 ± 5.98} & \multicolumn{3}{c}{81.62 ± 10.74} & \multicolumn{3}{c}{10.37 ± 11.64} \\
    \multicolumn{2}{c|}{SAM($X$)-box} & \multicolumn{3}{c}{\textbf{78.32 ± 9.81}} & \multicolumn{3}{c|}{\textbf{2.94 ± 1.48}} & \multicolumn{3}{c}{\textbf{90.74 ± 5.06}} & \multicolumn{3}{c}{\textbf{1.69 ± 3.02}} \\
    \bottomrule
    \end{tabular}    }%
  \label{table8}%
\end{table*}

\section{Discussion and Conclusion}
Our proposed SFDA method is model-agnostic as it does not rely on a specific network structure for segmentation. Instead, it focuses on data augmentation strategies and the use of SAM to obtain high-quality pseudo labels, which is combined with a reliability-aware loss function for learning in the target domain. As a result, it is more  general than previous SFDA methods that assume the source model has a specific network architecture~\citep{sun2020test,wu2023upl}. This framework is not limited to U-Net~\citep{ronneberger2015u} and can be adapted to other network structures, such as Transformer-based networks~\citep{chen2024transunet}, due to its focus on image adjustment and pseudo-label refinement rather than network design. By leveraging raw pseudo-labels obtained by the source model with T3IE, we create bounding box prompts that roughly identify location of the target object for SAM. The T3IE is elaborated to enhance the image contrast and reduce the gap between medical images and natural images where SAM was trained on, and the concatenation of the three outputs of T3IE further simulates RGB images that are more compatible for SAM inference.  
It is important to highlight that while $R^{\mathcal{H}_e}$, $R^{\gamma_D}$ and $R^{\gamma_S}$ are generated using SAM with the same bounding box prompt $\mathcal{B}$, the consistency region $\Omega _C$ remains relatively reliable and the impact of potential bias is minimized for two key reasons. First, the bounding box itself is derived from the enhancement and transformation processes of the input images $X^{\mathcal{H}_e}$, $X^{\gamma_D}$, and $X^{\gamma_S}$, which inherently introduces a degree of stability. Second, even if some regions are less reliable, the subsequent weighting process accounts for this by prioritizing regions where the three predictions align consistently. This approach leverages the diversity of the input images while focusing on the most consistent and reliable parts of the predictions, thereby enhancing the robustness of the final pseudo-labels.

This process enables SAM to produce more reliable pseudo-labels, which are then used to guide fine-tuning model parameters in the target domain without access to labeled images. 

Robust learning from noisy pseudo-labels is also essential in our method, as the refined pseudo-labels obtained by SAM may still have a low quality in some challenging cases. As the output of SAM is a mask without reliability information, we further leverage CMSO to obtain multiple outputs from SAM for reliability estimation. Compared with other reliability or uncertainty estimation methods that require a threshold to distinguish the reliable and unreliable regions in  pseudo-labels,  our CMSO based on consistency is parameter-free and easy to use.   
Finally, by effectively mining the reliable pseudo-labels for supervision and regularization on unreliable regions, our method ensures the robust adaptation of source model in the unannotated target domain without access to source images. 
It helps developing models that can adapt more efficiently to new domains,  enhancing the effectiveness and robustness of medical image segmentation models when deployed in clinical environment with unseen image distributions. 

There are areas in this work that still require improvement. Firstly, while it aligns with standard practices in the machine learning community, our method uses a labeled validation set in the target domain for model selection. In practice, this may not be available, making it more challenging for good adaptation.    Secondly, we only leveraged the zero-shot inference ability of SAM to refine pseudo-labels, while ignoring its strong feature representation ability, it would be of interest to distill the knowledge from SAM to the target-domain model for performance improvement. In addition, the setting of SFDA requires a set of target-domain images for training with multiple epochs, while in some cases these images may not arrive at once, therefore it is meaningful to explore test-time adaptation where target-domain images arrive in sequence in the future.  
Regarding the limitations of SAM, it currently supports only 2D images. However, recent work, such as Ma-SAM~\citep{chen2024ma} and Med-SAM~\citep{ma2024segment}, has explored the extension of SAM for 3D volume prediction. While this study primarily focused on 2D images, our method's core idea-using pseudo-labels as prompts and adapting large models-can be applied to 3D data as well. This opens up possibilities for applying our method to 3D medical image segmentation tasks in the future.

In conclusion, we propose a novel method based on SAM-Guided Reliable Pseudo-Labels  for Source-Free Domain Adaptation (SRPL-SFDA) in medical image segmentation tasks. Specifically, Test-Time Tri-branch Intensity Enhancement (T3IE) is proposed to mitigate the domain gap and enhance the quality of pseudo-labels that generate the bounding box prompt for SAM model. T3IE is also used to obtain SAM-compatible inputs that led to high-quality refined pseudo-labels in the target domain. 
A Consistency of Multiple SAM Outputs (CMSO) method based on T3IE is also proposed to obtain reliable regions of the pseudo-labels, which is then used for reliability aware pseudo-label supervision and regularization during model adaptation.  
Experimental results on two multi-domain medical image segmentation datasets demonstrate that our approach outperformed state-of-the-art source-free domain adaptation methods, and its performance is comparable to supervised training in the target domain, showing the potential of our method in obtaining high segmentation performance and robustness in unseen target domains.  
Furthermore, although our experiments were conducted on MRI datasets, the proposed method is not limited to MRI data and is applicable to other medical image modalities such as CT and PET. Our framework's model-agnostic nature allows it to be effectively applied to various imaging modalities, enhancing its utility in diverse clinical settings.

\section{Acknowledgment}
This work was supported by the National Natural Science Foundation of China under grant 62271115.
\bibliography{mybibfile}

\end{document}